\documentclass[10pt,a4paper,twocolumn]{article}
\usepackage[T1]{fontenc}
\usepackage[utf8]{inputenc}
\usepackage{lmodern,microtype}
\usepackage[a4paper,left=17mm,right=17mm,top=17mm,bottom=16mm,footskip=10mm]{geometry}
\usepackage{amsmath,amssymb,graphicx,booktabs,array,tabularx}
\usepackage[font=small,labelfont=bf,labelsep=colon]{caption}
\usepackage{titlesec}
\usepackage[hidelinks]{hyperref}
\hypersetup{pdftitle={Borrowing Shapes Across Instruments: Synthetic Training Images for Solar Filament Instance Segmentation},pdfauthor={Ikjoo Oh, Taegwon Kang, Seohee Min}}
\titleformat{\section}{\large\bfseries}{\thesection}{0.7em}{}
\titleformat{\subsection}{\normalsize\bfseries}{\thesubsection}{0.75em}{}
\titlespacing*{\section}{0pt}{1.3ex plus .4ex minus .2ex}{.8ex}
\titlespacing*{\subsection}{0pt}{1.15ex plus .4ex minus .2ex}{.6ex}

\begin{document}
\twocolumn[{%
\begin{minipage}{\textwidth}
\vspace*{3pt}
\begin{center}
{\Large\bfseries Borrowing Shapes Across Instruments: Synthetic Training Images for\\Solar Filament Instance Segmentation\par}
\vspace{6pt}
{\normalsize Ikjoo Oh, Taegwon Kang, Seohee Min\par}
\vspace{2pt}
{\small University of Wisconsin--Madison\par}
\end{center}
\vspace{5pt}
\noindent\textbf{Abstract.}
Training images assembled from external solar-filament masks must reconcile the morphology of the donor with the appearance and annotation conventions of the target observations.
We examine two construction procedures for full-disk GONG H$\alpha$ segmentation: copy-paste of annotated GONG cutouts and rendering of CHASE mask geometry with attenuation statistics estimated from GONG training observations.
The latter procedure separates the transferred outline from the donor intensity and preserves the aligned target background outside inserted masks.
Early YOLO experiments recorded internal validation PQ of 0.39 with real GONG observations alone and 0.43 with GONG copy-paste; these historical runs are distinct from the fixed-development comparisons.
Evaluation uses 91 GONG development images with 101 annotator--image groups and 708 reference instances.
At a common foreground threshold and minimum component area, a ConvNeXt-Tiny U-Net trained with CHASE-shape composites obtains panoptic quality (PQ) 0.431117, compared with 0.429419 for the GONG-copy-paste configuration.
The difference of 0.001698 accompanies 6.65\% fewer false positives and 3.04\% more misses, while mean matched-mask IoU is nearly unchanged.
Selecting a separate decoding threshold gives PQ 0.431339 for the shape model.
In a distinct continuation experiment, training both branches of a GONG ensemble on CHASE observations alone reduces GONG PQ from 0.471930 to 0.106709 under the retained fusion rules.
The full YOLO + U-Net pipeline trained with original GONG observations, GONG copy-paste, and CHASE-shape composites achieves PQ 0.4819 on the same 91 development frames, compared with 0.4719 for the original ensemble.
The configurations differ in synthetic sampling, optimization history, decoding, and execution environment, so these results do not isolate a causal advantage of one synthesis method.
They establish a feasible route for using external mask geometry and a higher recorded full-pipeline development score; the standalone U-Net difference remains a small, selection-dependent precision--recall shift, rather than demonstrated improvement in boundary fidelity or independently validated generalization.\par

\smallskip
\noindent\textbf{Keywords.} Solar filaments; H$\alpha$ observations; synthetic data; instance segmentation; domain shift\par

\end{minipage}
\vspace{3pt}
}]
\section{Introduction}

Solar filaments are observed in absorption against the solar disk in H$\alpha$ and contain cool plasma suspended in the solar atmosphere. Extracting their spatial extent from ground-based archives supports studies of filament morphology and solar activity, but faint boundaries, nearby structures, and varying image quality complicate automated measurement~\cite{r1}. A useful instance catalog must identify complete objects as well as foreground pixels. A mask that captures only a dark core can be correctly located yet fail an instance-overlap criterion, while a connected foreground region can merge neighboring filaments.

MAGFiLO provides manually annotated filaments in Global Oscillation Network Group (GONG) observations~\cite{r2}, enabling an instance-segmentation evaluation in the Solar Filament Segmentation Challenge 2026~\cite{r3}. Additional H$\alpha$ archives offer more examples of morphology, but their intensities and annotations are not interchangeable. Filter characteristics, acquisition geometry, image processing, and label construction can change a structure's apparent width and fragmentation. Directly reusing an external image therefore couples additional shape variation with a shift in appearance and supervision.

Synthetic augmentation allows these factors to be treated separately. Within one observing system, an annotated cutout transfers both a filament outline and measured grayscale texture. Across instruments, a binary mask can instead supply geometry while a renderer determines the appearance in a target background. Copy-paste has been studied as a practical augmentation for natural-image instance segmentation~\cite{r4}. Its application to solar images additionally requires control of disk placement, contrast, boundary labels, and the provenance of both donor and background.

We study these two construction routes using GONG cutouts and masks from the Chinese H$\alpha$ Solar Explorer (CHASE)~\cite{r5}. The CHASE route uses automatic semantic maps from AHAS~\cite{r6,r7}, extracts components as geometry donors, and estimates their attenuation from GONG training observations. CHASE grayscale intensities are not copied into the synthetic training images. This differs from translating a complete external observation into the target instrument's appearance, and it does not require external observations at GONG inference.

Our contribution is an explicit description of the synthesis procedure, a development comparison of two native U-Net training configurations, a separate analysis of CHASE-only continuation in a detector--semantic ensemble, and a full-pipeline evaluation using both synthetic sources. We also retain the historical within-GONG YOLO comparison at its recorded precision. We examine which changes reflect object selection and which reflect overlap of matched masks. The study is exploratory: it does not provide a matched real-only control or hold all training conditions fixed across synthesis sources. We therefore distinguish a trainable construction and its observed operating point from a demonstrated augmentation benefit.

\section{Related work}

\subsection{Solar filament detection across observations}

Diercke et al.~\cite{r1} combine manual ChroTel bounding boxes, YOLOv5 detection, and automatic mask generation to train a U-Net on an expanded ChroTel/GONG collection. Their work illustrates how imperfect labels and image correction can support large-scale filament extraction, with qualitative application to KSO observations. In that method, the detector helps generate segmentation supervision. In our synthesis procedure, an external binary map supplies an outline that is inserted at a chosen location in a GONG background; a target-derived statistical profile supplies the intensity decrement. These are different uses of auxiliary observations.

The distinction also extends to evaluation. Diercke et al.~\cite{r1} report a detection-oriented comparison using boxes derived from segmentation maps, including margin expansion and a relaxed bbox matching threshold. Our experiments use pixel-mask IoU and unmatched-instance penalties. We do not compare their reported detection percentages numerically with the PQ values below. Hu et al.~\cite{r6} develop MORDEN and the AHAS/MHAS data resources for filament semantic segmentation. Their automatically generated AHAS maps provide useful geometry, but connected components of a semantic map are not verified physical instance identities.

\subsection{Appearance transfer and compositional augmentation}

Ghiasi et al.~\cite{r4} show that simple instance copy-paste can be effective in natural-image segmentation. In solar imaging, the appearance of an inserted structure must also be compatible with the target disk and local background. Jarolim et al.~\cite{r8} study instrument-to-instrument translation of solar observations using deep learning. Our renderer addresses a narrower task: it constructs an empirical absorption pattern inside a specified binary support rather than recovering a calibrated observation from another instrument. It is neither a learned image-translation network nor a radiative-transfer inversion.

The segmentation models use established U-Net~\cite{r9} and ConvNeXt~\cite{r10} components. The methodological question is how synthetic image--mask pairs interact with the target task, rather than whether a new backbone is superior. Panoptic quality~\cite{r11} is useful here because it combines matched-mask overlap with missed and spurious objects. A gain in foreground plausibility or precision need not imply a gain in recall, and a small PQ increase need not indicate better boundaries.

A physics-based simulator such as FILOS computes H$\alpha$ absorption from a three-dimensional filament model and applies an instrumental response~\cite{r17}. Our construction instead uses a library of measured target-image profiles. It can generate training pairs from existing donor masks, annotated target observations, and a suitable background without constructing a three-dimensional filament model or simulating the telescope response.

\section{Observations and experimental units}

\subsection{Target data and physical-image grouping}

The target images are $2048 \times 2048$ grayscale full-disk GONG observations. The fixed split contains 616 physical training frames with 1053 annotator--image records and 7491 filament annotations, and 91 development frames with 101 records and 708 reference instances. A physical frame may have more than one annotation variant. All variants of the same filename are grouped in one partition; predictions are later evaluated separately against each available annotation group. Filament chirality is not a prediction target in this study, and all supplied filament categories contribute to foreground supervision.

The historical synthesis metadata initially describes 924 donor-side and 230 validation-side annotation identifiers. Although those identifiers are disjoint, their physical filenames overlap in 115 cases. Grouping every annotation variant of a donor-side filename into training yields the fixed 616/91 physical-image split used for the native experiments. This prevents direct filename leakage but does not establish temporal independence: thirteen observing dates contain frames in both partitions. Observations from different sites can depict closely related solar structures. The 91-frame cohort is reused for checkpoint and decoding selection and is therefore a development set.

\subsection{External masks and background assumptions}

The CHASE pool comprises 695 eligible 2024--2025 observations from the AHAS release~\cite{r6,r7}, excluding the older observations shared with MHAS. The masks are automatic binary semantic maps, not manually verified instance partitions. An eight-connected component can join touching filaments or represent only a fragment of one structure. We retain this distinction when using components as shape donors and as proxy polygons in the continuation experiment. Explicit exclusions and unobserved support are respected. Their observing dates follow those in the GONG development split.

Both synthesis routes use the GONG Learmonth frame acquired at 23:59:30 UT on 18 June 2020 at 6562.8~\AA{}~\cite{r15}. The frame was selected using a same-day Prairie View observing log that reported no disk filaments earlier in the day~\cite{r16}; its date is absent from the 707-frame GONG collection. We treat the later exposure as filament-free for label generation. A working copy is converted to grayscale and aligned to the 2048-pixel grid and reference disk. The observing log and renderer quality checks do not certify that every faint feature is absent. Any residual unlabeled structure is repeated in every synthetic variant, including compositions with no inserted objects. The use of one background also limits texture diversity. The procedural illustration in Figure~\ref{fig:2} is separate from this experimental background.

\begin{figure*}[t]
\centering
\includegraphics[width=\textwidth]{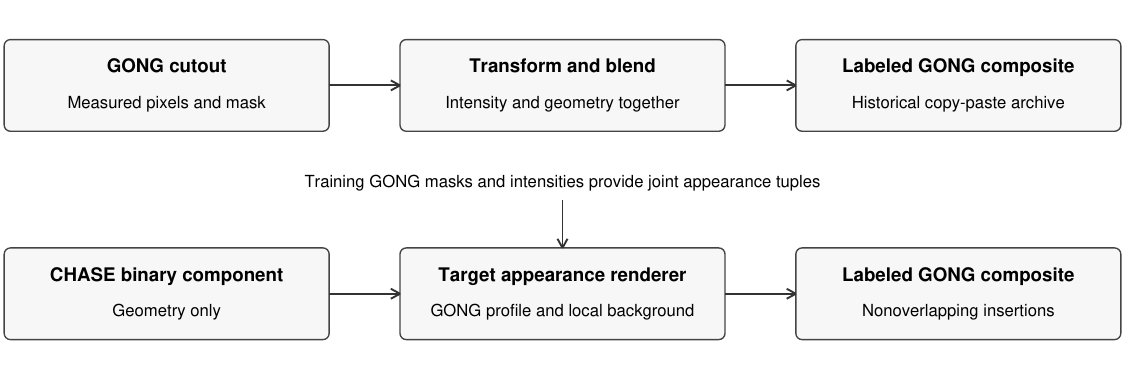}
\caption{The two construction routes. GONG copy-paste carries measured donor pixels and a mask into a common background. CHASE-shape synthesis carries only mask geometry; a joint appearance tuple estimated from GONG training objects controls the rendering. Both routes generate labeled training inputs before the shared GONG preprocessing. The diagram represents data construction, not the two branches of the inference ensemble.}
\label{fig:1}
\end{figure*}

\subsection{Temporal-block cross-validation protocol}

To assess baseline performance across observing periods, five-fold cross-validation used 707 distinct GONG frames with real observations for training in each fold. Observations were assigned to folds in 27-day blocks, with all annotation variants of a frame kept together. Each frame was held out once, with 140--143 development frames per fold. Training excluded observations within three days of a validation date. This protocol is distinct from the fixed 91-frame cohort used for the synthesis and continuation results below; those results are not reported as five-fold performance.

\section{Construction of synthetic images}
\subsection{Copy-paste of GONG cutouts}
An annotated GONG filament supplies a cropped grayscale patch and binary mask. For each composition, the requested number of objects is sampled from the positive-instance-count distribution of donor annotation records. Cutouts are selected uniformly, rotated within $[-25^{\circ},25^{\circ}]$, scaled within $[0.85,1.20]$, and reflected horizontally with probability 0.5. Masks must fit inside a disk of radius 900 pixels centered on the native grid. The historical generator permits an insertion to overlap previously placed masks by up to 10\% of its area.

Donor intensities are multiplied by a factor sampled from $[0.90,1.10]$ and blended into the background using a $5\times5$ Gaussian-smoothed mask. The transformed binary support supplies the annotation contour. The existing collection contains 462 compositions and 3427 instance annotations, including ten background-only images. Positive compositions contain between one and 24 instances. It is a fixed synthetic archive, not a new set regenerated during every epoch. The composition count is half of the 924 donor-side annotation records, which refer to 616 physical training frames; all 91 fixed-development frames lie outside that donor pool.

Reproduction must preserve the historical label convention. Legacy synthetic polygons are rasterized using coordinate rounding followed by OpenCV \texttt{fillPoly}, whereas real target annotations use the COCO-compatible path. An archive audit found that replacing the legacy convention with COCO polygon rasterization reduces mean mask area by about 10\%. That discrepancy would alter training labels rather than implement a benign format conversion. The contour and compositing conventions are consequently part of the synthesis specification~\cite{r14}.

\subsection{Selection and transformation of CHASE geometry}
CHASE image/mask grids are reconciled using the source release convention before disk alignment. In the verified 2025 exports, the image dimensions differ from the 2048-pixel mask grid; images are resized to that grid before the common alignment, while masks retain categorical values and use nearest-neighbor resampling. Components touching excluded or unobserved support are not used. A donor observation is sampled before its component, limiting overrepresentation of frames containing many small fragments.

Within the chosen observation, component selection favors areas close to a randomly selected GONG training-object profile. Shapes receive a random quarter-turn, optional horizontal reflection, and isotropic scale in $[0.8,1.25]$. Each of 616 synthetic variants contains three nonoverlapping insertions, for 1848 generated instance masks. All inserted support remains within 94\% of the disk radius; an eight-pixel placement margin separates insertions. The fixed density of three objects differs from the empirical-count copy-paste archive and is an additional experimental factor.

\begin{figure*}[t]
\centering
\includegraphics[width=\textwidth]{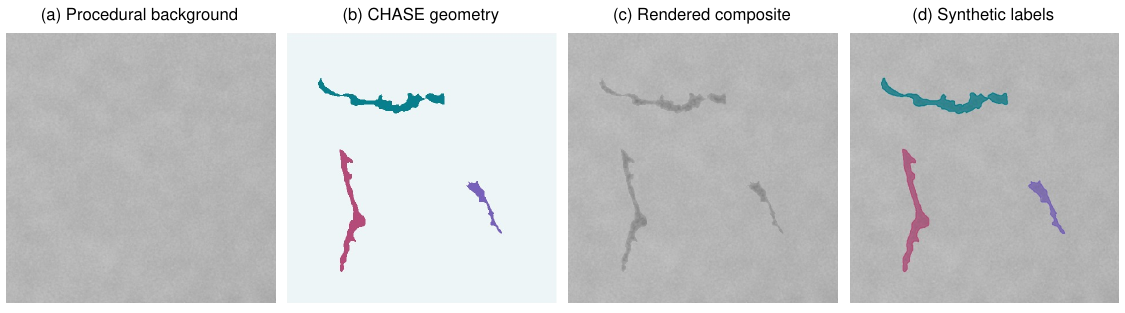}
\caption{Illustration of the CHASE-shape renderer using a procedural grayscale background and three public AHAS components. Panels show (a) background, (b) transferred geometry, (c) composite, and (d) generated instance labels. Fixed illustrative parameters are $\tau_c=0.25$, $\rho_e=0.3$, $v=0.15$, and $w=5$; these are not measured GONG calibration profiles, and this is not a sample from the training synthesis archive. Geometry is translated from observation \texttt{RSM20240101T061747\_0000\_HA} in the Hu et al. dataset~\cite{r7}, CC BY 4.0. Display colors identify synthetic labels, not prediction accuracy. Adapted from the reproducible project renderer demonstration~\cite{r14}.}
\label{fig:2}
\end{figure*}

\subsection{Estimating target appearance}
Appearance calibration uses GONG training observations only. For each physical image, one annotation variant is selected deterministically, and up to six of its objects are considered. The union of known foreground from all available annotators protects the surrounding ring. The run description reports 2993 accepted profiles; this count represents a sampled and quality-filtered calibration set, not every annotated training object.

For an object mask, a local plane is fitted to a surrounding ring after excluding nearby mask pixels, known foreground, and saturated or near-zero intensities. At native resolution, the outer dilation uses a 16-pixel margin and the immediate $3\times3$ dilation is excluded. The plane fit uses three iterations of residual rejection based on a scaled median absolute deviation. With estimated background $\widehat{B}$ and observed image $I$, the empirical decrement is
\begin{equation}
 a(x)=\log\!\left(\frac{\widehat{B}(x)+1}{I(x)+1}\right).
 \label{eq:decrement}
\end{equation}
The core comprises pixels above the 60th percentile of interior boundary distance; the edge comprises pixels below the 25th percentile. Each accepted tuple records median core attenuation, edge-to-core ratio, robust core variation, local background intensity, normalized disk radius, area, and a width estimate. Profiles without a reliable dark decrement or trustworthy surrounding plane are rejected. The decrement is an empirical image parameter and is not interpreted as a calibrated optical depth.

At an insertion site with background brightness $b$, radius $r$, and shape area $A$, the distance to profile $j$ is
\begin{equation}
 D_j=\left(\frac{b-b_j}{40}\right)^2+
     \left(\frac{r-r_j}{0.25}\right)^2+
     \left(\frac{\log(A/A_j)}{2}\right)^2 .
 \label{eq:distance}
\end{equation}
A tuple is sampled among the twelve nearest profiles with probability proportional to $\exp[-0.5(D_j-D_{\min})]$. Joint tuple sampling retains the measured association between attenuation, edge ratio, and internal variation. It does not independently draw these values from unrelated marginal distributions.

\subsection{Rendering and label preservation}
\begin{figure*}[tp]
\centering
\includegraphics[width=.94\textwidth]{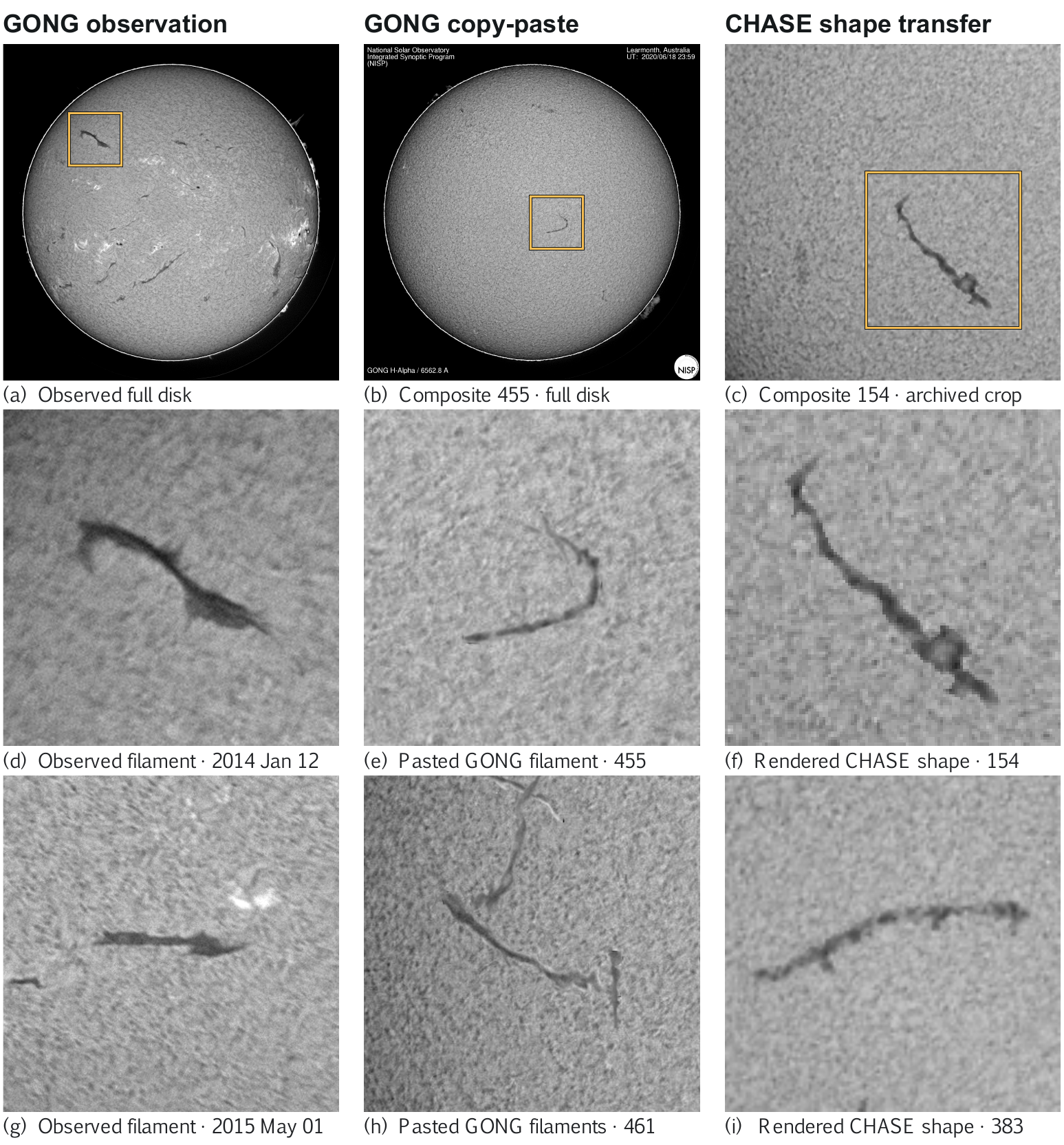}
\caption{Measured GONG H$\alpha$ filaments (left), GONG copy-paste composites (middle), and CHASE shapes rendered on GONG (right). Examples illustrate visible morphology. Boxes in (a)--(c) locate the middle-row details. Panels (a) and (b) show full disks; (c) is an archived regional preview. Middle- and bottom-row fields span approximately 320 and 480 original image pixels per side. CHASE details use unmarked composite panels of downsampled run previews. All panels use the same linear grayscale display window, 55--180. GONG observations and annotations are from MAGFiLO~\cite{r2}, the background is from GONG Learmonth~\cite{r15}, and CHASE masks derive from AHAS~\cite{r6,r7}.}
\label{fig:3}
\end{figure*}

Let $d(x)$ be the distance inside a transformed mask, normalized by its 90th percentile and clipped to $[0,1]$. For sampled core attenuation $\tau_c$, edge ratio $\rho_e$, relative variation $v$, and smoothed standardized noise $\eta(x)$, the rendered field is
\begin{equation}
 \begin{aligned}
  \tau(x) &= \tau_c[\rho_e+(1-\rho_e)d(x)^{0.7}]\\
  &\quad\times\max[0.2,1+v\eta(x)].
 \end{aligned}
 \label{eq:attenuation}
\end{equation}
Noise is Gaussian-smoothed with scale $\max(0.7,w/4)$, where $w$ is the sampled width estimate, standardized over mask support, and clipped to $[-2,2]$. Within the support, the composite is
\begin{equation}
 \begin{aligned}
 J(x)&=(B(x)+1)e^{-\tau(x)}-1,\\
 I_{\mathrm{syn}}(x)&=\operatorname{clip}_{[0,255]}\{\operatorname{round}[J(x)]\}.
 \end{aligned}
 \label{eq:composite}
\end{equation}
Outside inserted support, raw aligned-background pixels remain unchanged. The generator requires at least 90\% of each mask to darken by one or more grayscale levels, a median decrement of at least one level, and no zero-valued inserted pixels. It rejects clipping, intersections with reserved support, and selected unexplained strong dark features at the insertion site. These are numerical visibility and placement checks, not validation of physical realism. Exact instance RLEs are generated for the inserted shapes. Shared limb correction is applied afterward; the outside-mask preservation assertion refers to the raw aligned composite before that correction. Figure~\ref{fig:3} compares measured GONG structures with archived GONG-copy-paste and CHASE-shape composites.

\begin{table*}[t]
\centering
\caption{Native training configurations. Both runs complete 100 epochs, but their selected checkpoints and execution environments differ. The first and last real-only intervals are a data schedule, not a learning-rate warmup.}
\label{tab:1}
\small
\renewcommand{\arraystretch}{1.12}
\setlength{\tabcolsep}{4pt}
\begin{tabularx}{\linewidth}{@{}p{0.33\linewidth}XX@{}}
\toprule
Setting & GONG copy-paste & CHASE shapes \\
\midrule
Encoder and decoder & ConvNeXt-Tiny U-Net & Same \\
Initialization & ImageNet encoder and new decoder & Same \\
Crop and presentations per epoch & $1024\times1024$ and 400 & Same \\
Epochs and selected checkpoint & 100 and epoch 84 & 100 and epoch 82 \\
Synthetic probability in epochs 4--94 & 0.20 & 0.50 \\
Real-only epochs & 1--3 and 95--100 & Same \\
Optimizer and initial LR & AdamW and 0.0001 & Same \\
Weight decay and clipping & 0.01 and norm 1.0 & Same \\
Cosine schedule and minimum LR & 100 epochs and 0.000005 & Same \\
Batch and accumulation & 2 and 2 & 1 and 4 \\
GPU and precision & A100 40 GB and BF16 & T4 and FP16 \\
Random seed & 42 & 42 \\
\bottomrule
\end{tabularx}
\end{table*}

\section{Training and evaluation}
\subsection{Native segmentation configurations}
The native branch uses a segmentation-models-pytorch U-Net with a ConvNeXt-Tiny encoder~\cite{r9,r10,r13}. Both compared runs initialize the encoder from ImageNet and the decoder from new weights; neither starts from a previous solar-trained U-Net. Independent foreground and auxiliary spine logits are produced. Foreground supervision is the union of the selected annotator's masks. Foreground BCE and soft Dice are combined with weight 0.1 on balanced spine BCE where genuine spine supervision exists. Missing synthetic spines are excluded from that term rather than treated as negative labels. The foreground output alone supplies the evaluated masks.

An epoch comprises 400 image-first random $1024\times1024$ crop presentations. Sampling probabilities are 0.50 for a small positive object, 0.25 for a general positive object, and 0.25 for a uniform crop, with fallbacks for empty requested pools. Small objects are defined by the training-only first area quartile. Crop and annotation sampling do not increase the number of independent observations. D4 transforms, gamma in $[0.9,1.1]$, and brightness multiplication in $[0.85,1.15]$ provide augmentation.

Both runs use an effective batch size of four and 100 nominal optimizer updates per epoch; mixed-precision overflow can affect realized updates. During epochs 4--94, the original configuration samples the GONG archive with probability 0.20, while the shape configuration samples CHASE composites with probability 0.50. Epochs 1--3 and 95--100 use real GONG only. These differences in source, density, mixture, batch, precision, and hardware prevent attribution to transferred geometry alone. The baseline checkpoint is evaluated in the T4 comparison environment without being used to initialize the new run.

\subsection{Image correction and decoding}
GONG images receive the existing radial limb correction. A fixed disk centered at $(1024,1024)$ with radius 900 pixels is divided into 100 radial bins. Bin medians are fitted with a cubic smoothing spline, and a central reference within normalized radius 0.15 sets a gain clipped to $[0.25,4]$. Corrected values are clipped to $[0,255]$, and off-disk pixels are zero. A grayscale image is repeated into three channels and receives ImageNet normalization; no additional color measurement is introduced.

Native inference uses nine 1024-pixel tiles with origins 0, 768, and 1024 on each axis. The nominal overlap is 25\%; endpoint coverage makes the final spacing nonuniform. Probabilities are combined using a separable Hann window with minimum weight 0.001 and division by the accumulated window. Foreground thresholding is followed by eight-connected-component filtering in native pixel coordinates. Components are operational instances for this evaluation and can differ from physical filament identities. The native-only comparison uses this single-orientation tiled decoding, distinct from the multiview ensemble below.

Foreground thresholds 0.35, 0.50, and 0.65 and minimum areas 100, 200, and 300 pixels define the development grid. The CHASE-shape run is evaluated at epoch 1 and every two epochs through epoch 100, yielding 51 saved evaluation points. We report its selected epoch both at the baseline decoder, threshold 0.50 and area 300, and at the selected threshold 0.35 with the same area. Using a common decoder removes one difference but does not remove checkpoint-selection bias.

\subsection{Separate CHASE-only continuation}
The continuation study starts from the existing GONG-trained YOLO26s-seg~\cite{r12} and the original native checkpoint. Each branch is continued for 60 epochs on the 695 aligned CHASE observations with no GONG replay or final GONG calibration stage. The detector sees each full 2048-pixel frame once per epoch and receives component-derived proxy polygons. The native branch sees one 1024-pixel crop per observation per epoch and receives foreground supervision only; this visits every observation but does not cover every pixel each epoch. Initial learning rate is 0.00002, with AdamW, weight decay 0.01, cosine decay over 60 epochs, and seed 42. CHASE inputs receive disk alignment without the additional GONG contrast correction or the photometric augmentation of the mixed native runs.

Four original/continued checkpoint combinations are compared on the same GONG development frames. Both continued weights are taken from epoch 60. Evaluation retains GONG limb correction and the original parallel ensemble rules: detector geometric, photometric, and tiled views supply proposals, while native tiled foreground and four-orientation TTA supply support. The frozen fusion performs conditional additions, guarded boundary refinement, weak-mask pruning, and disjoint pixel assignment~\cite{r14}. Its 30 detector views and 36 native tile evaluations differ from native-only decoding. The comparison measures the consequences of continuation under retained rules; it does not isolate representation change from output calibration.

\subsection{Metric and uncertainty}
Following the challenge-style implementation, every mask pair with strict IoU greater than 0.5 belongs to the qualifying set $M$. A false positive has no qualifying reference and a false negative has no qualifying prediction. IoUs and counts are pooled across annotator--image groups before computing
\begin{equation}
 \mathrm{PQ}=\frac{\sum_{(p,g)\in M}\mathrm{IoU}(p,g)}
 {|M|+\tfrac12\mathrm{FP}+\tfrac12\mathrm{FN}}.
 \label{eq:pq}
\end{equation}
We retain all qualifying pairs rather than adding an assignment algorithm. For overlapping diagnostic outputs, their number can differ from the count of uniquely matched objects; mask validity and matching multiplicity must therefore be checked separately. For the disjoint decoded native outputs, precision and recall are calculated from the reported counts, and segmentation quality (SQ) is the mean qualifying IoU. The evaluator discovers groups from GT instance rows and cannot characterize entirely blank reference groups omitted by that representation. Scores are not averages of per-image PQ.

All four archived full-ensemble continuation prediction files were rescored. The paired physical-frame bootstrap samples 91 frames with replacement 20000 times, retaining their annotation variants and recomputing pooled PQ for both fixed models. The reported 95\% interval is conditional on these models and treats frames as exchangeable. It does not include training-seed variation, temporal dependence, or model-selection uncertainty. Paired baseline masks are unavailable for the native synthesis comparison, so we do not report a newly computed paired interval for its small PQ difference. This continuation interval is not an uncertainty estimate for the augmented full-pipeline gain. Historical YOLO scores are retained separately at two-decimal precision, with their split and aggregation limitations.

\subsection{Historical and combined-source configurations}
The early GONG copy-paste experiment compared YOLO segmentation runs using real observations alone and real observations with synthetic GONG cutouts. An archived notebook contains alternative split-construction branches and a per-record summary, so notebook defaults alone do not establish the exact aggregation for those two runs. Their internal validation results are reported separately from the fixed-cohort evaluation.

The combined-source experiment trains the full YOLO + U-Net pipeline using original GONG observations, GONG copy-paste composites, and CHASE-shape composites rendered on GONG. It is evaluated on the same 91 development frames as the original GONG ensemble. The augmented-pipeline result is reported at threshold 0.50, whereas the original ensemble retains its multiple-threshold fusion policy. Standalone U-Net settings in Table~\ref{tab:1} must not be assumed to specify every detail of this separate full-pipeline training configuration.

\section{Results}

\subsection{Native comparison at a common decoder}

At threshold 0.50 and minimum area 300, CHASE-shape training increases PQ by 0.001698, from
0.429419 to 0.431117. The endpoint counts contain 21 fewer false positives and seven fewer
matches, with seven additional misses. Relative to the baseline, FP decreases by 6.6\% and FN
increases by 3.0\%. Precision increases by about 1.29 percentage points and recall decreases by
about 0.99 percentage points. These are differences in aggregate counts, not a claim that 21
corresponding false detections were individually removed. Establishing per-object gains and
losses requires both prediction sets.

Mean matched IoU is 0.674674 for the baseline and 0.674593 for the shape model. Their near
equality indicates that the observed PQ increase is associated with the balance of matched and
unmatched objects rather than improved overlap among matched masks. Recognition quality
calculated from the counts increases from approximately 0.636485 to 0.639077. The result is
therefore best described as a modest shift toward more conservative instance selection at the
recorded operating point, not evidence of higher boundary fidelity.

\begin{table*}[t]
\centering
\caption{Native-only development results with minimum area 300 pixels. P and R are precision and recall calculated from the reported counts. The first two rows use a common threshold; the final row changes the shape model's threshold after development selection.}
\label{tab:2}
\small
\renewcommand{\arraystretch}{1.12}
\setlength{\tabcolsep}{4pt}
\begin{tabularx}{\linewidth}{@{}Xrrrrrrr@{}}
\toprule
Training configuration & Threshold & PQ & TP & FP & FN & P \% & R \% \\
\midrule
GONG copy-paste 20\%, epoch 84 & 0.50 & 0.429419 & 478 & 316 & 230 & 60.20 & 67.51 \\
CHASE shapes 50\%, epoch 82 & 0.50 & 0.431117 & 471 & 295 & 237 & 61.49 & 66.53 \\
CHASE shapes 50\%, selected decoder & 0.35 & 0.431339 & 472 & 295 & 236 & 61.54 & 66.67 \\
\bottomrule
\end{tabularx}
\end{table*}

\subsection{Decoder and checkpoint selection}

Selecting threshold 0.35 for the shape model gives PQ 0.431339, with TP 472, FP 295, and FN 236.
Relative to the common-decoder result, the aggregate table gains one match with no increase in
FP. Relative to the baseline operating point, the PQ difference is 0.001920. Decoder selection
is part of this latter difference; it should not be presented as a fixed-decoder estimate of the
synthesis effect.

The selected shape checkpoint is epoch 82. Its final epoch-100 model reaches PQ 0.420170 at its
best grid setting, and only the selected epoch exceeds the baseline score among the 51 recorded
evaluation points. This sensitivity to checkpoint selection and the absence of repeated training
seeds limit an interpretation of the small difference as a stable improvement. The comparison
also changes the synthetic probability from 0.20 to 0.50 and the execution environment. We do
not infer a statistically established advantage from either selected point.

\subsection{Direct continuation under fixed fusion}

\begin{table}[t]
\centering
\caption{Separate full-ensemble continuation comparison reported on the fixed GONG development cohort. O denotes original GONG weights and C denotes the corresponding CHASE-continued epoch-60 weights. These values use full fusion, not the decoder of Table~\ref{tab:2}.}
\label{tab:3}
\small
\renewcommand{\arraystretch}{1.12}
\setlength{\tabcolsep}{4pt}
\begin{tabularx}{\linewidth}{@{}llrrrrX@{}}
\toprule
YOLO & U-Net & PQ & TP & FP & FN & \\
\midrule
O & O & 0.471930 & 487 & 204 & 221 & \\
C & O & 0.245581 & 289 & 531 & 419 & \\
O & C & 0.319846 & 219 & 46 & 489 & \\
C & C & 0.106709 & 101 & 367 & 607 & \\
\bottomrule
\end{tabularx}
\end{table}

Replacing both branches lowers PQ from 0.471930 to 0.106709, a difference of -0.3652 at the
reported precision. The recorded paired frame-bootstrap interval for this change is [-0.3985,
-0.3296]. Matched-reference coverage in the reported counts falls from 487/708 (68.8\%) to
101/708 (14.3\%). The branch substitutions have different effects: the continued detector with
the original native model gives 531 FP, whereas the original detector with the continued native
model retains 219 matches and produces 489 FN.

\begin{figure*}[tp]
\centering
\includegraphics[width=\textwidth]{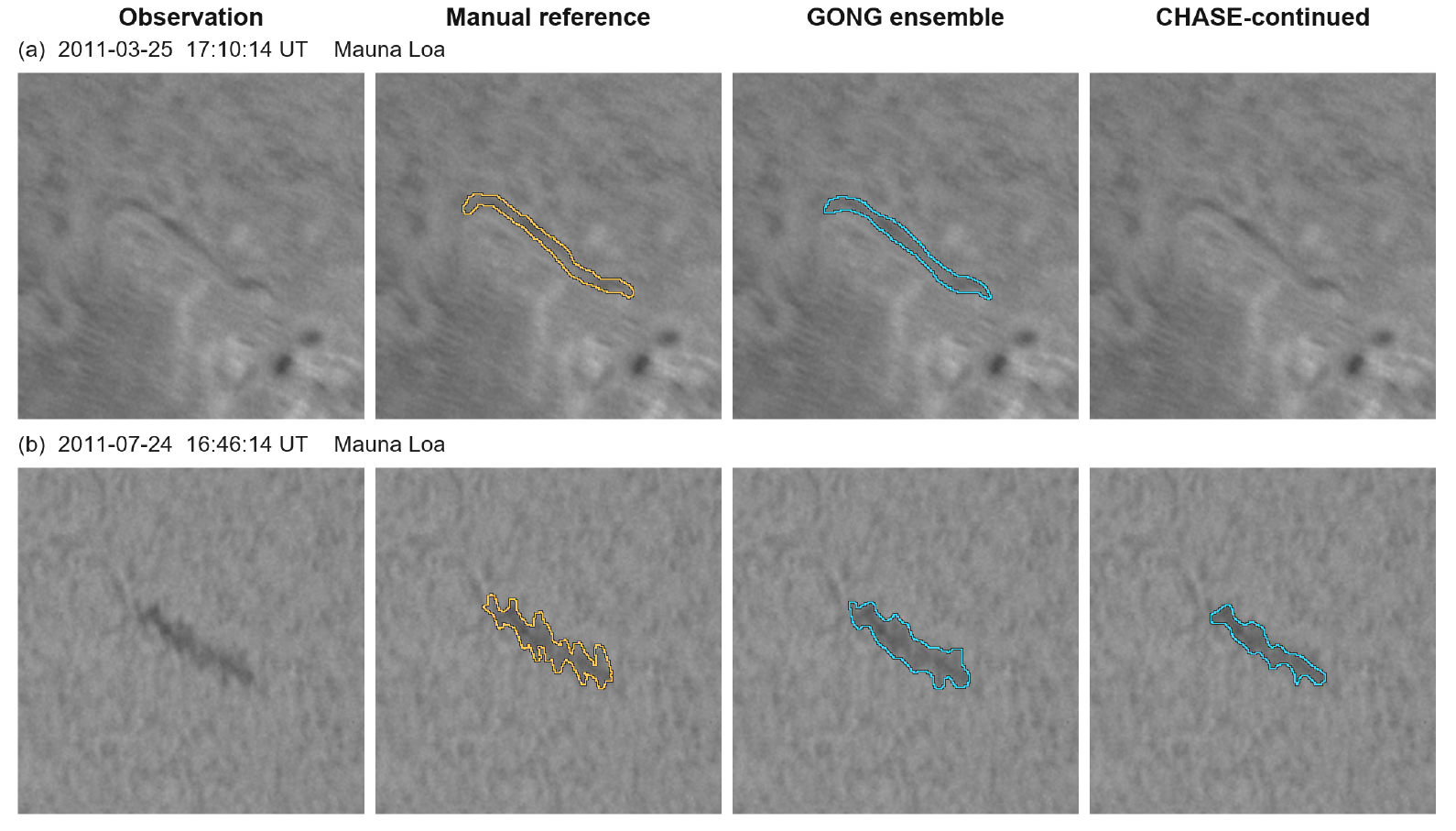}
\caption{Archived predictions for two $240\times240$-pixel crops from held-out GONG observations. Gold outlines trace MAGFiLO reference instances~\cite{r2}; cyan outlines show ensemble predictions. CHASE-continued branches use the original fusion rules. After continuation no mask is produced in crop (a), while both ensembles recover the filament in crop (b). These selected examples illustrate behavior rather than estimate its frequency.}
\label{fig:4}
\end{figure*}

These are outputs of the fused system. They are not direct measurements of the individual
branches' intrinsic quality and do not demonstrate that CHASE morphology is uninformative.
Additional training on CHASE alone, proxy-instance granularity, appearance shift, and the
retained GONG-tuned selection rules all intervene. In particular, changes in native support can
cause the same detector candidates to be rejected. The large decline identifies a failure of
this continuation procedure on GONG, while the responsible factors remain confounded.
Figure~\ref{fig:4} shows archived reference-aligned predictions for two held-out GONG crops.

\subsection{Training exposure at completion and selection}

Across the complete 100-epoch schedules, both runs present 40000 crops. Reconstructing the
domain draws from the declared seed and schedule gives 32569 real-GONG presentations for the
original configuration and 21549 for the shape configuration. The reduction of 11020
presentations is about 33.8\% of the former total. These are descriptive exposure statistics for
completed schedules, not exposures of the checkpoints compared in Table~\ref{tab:2}.

Stopping the same reconstruction at epochs 84 and 82 yields the distinct totals in
Table~\ref{tab:4}. This calculation uses the sampler's deterministic first draw for each epoch
and crop index; it does not verify whether every attempted update was executed. Moreover, both
configurations depend on the same annotated GONG training pool, and the renderer uses GONG
annotations for appearance calibration. Fewer repeated real crops therefore do not establish a
reduction in annotation acquisition, unique target images, GPU time, or energy. Hardware and
precision differ between the runs, and total synthesis and preprocessing costs have not been
measured under a common runtime.

\begin{table*}[t]
\centering
\caption{Deterministic reconstruction of crop-domain draws from the implemented seed-42 sampler, not a new recovery of executed training logs. Full-schedule totals and selected-checkpoint totals refer to different model states.}
\label{tab:4}
\small
\renewcommand{\arraystretch}{1.12}
\setlength{\tabcolsep}{4pt}
\begin{tabularx}{\linewidth}{@{}Xrrr@{}}
\toprule
Configuration and endpoint & Total crops & Real GONG & Synthetic \\
\midrule
Copy-paste, full epoch 100 & 40000 & 32569 & 7431 \\
CHASE shape, full epoch 100 & 40000 & 21549 & 18451 \\
Copy-paste, selected epoch 84 & 33600 & 26952 & 6648 \\
CHASE shape, selected epoch 82 & 32800 & 16725 & 16075 \\
\bottomrule
\end{tabularx}
\end{table*}

\subsection{Historical within-GONG copy-paste results}

In the early YOLO experiment, internal validation PQ was 0.39 with original GONG observations
and 0.43 with copy-paste augmentation, a difference of 0.04 at the recorded two-decimal
precision (Table~\ref{tab:5}). The mixed model scored 0.37 on the public leaderboard; a paired
real-only leaderboard result is unavailable. These aggregate scores do not identify how much of
the difference comes from FP, FN, or mask overlap, and their historical split and aggregation
limitations prevent treating them as a controlled estimate on the later fixed cohort.

\begin{table}[t]
\centering
\caption{Historical early-YOLO internal validation results, recorded to two decimal places. These runs are distinct from the fixed 91-frame comparison.}
\label{tab:5}
\small
\renewcommand{\arraystretch}{1.12}
\setlength{\tabcolsep}{4pt}
\begin{tabularx}{\linewidth}{@{}Xr@{}}
\toprule
Early YOLO training inputs & Validation PQ \\
\midrule
Original GONG observations & 0.39 \\
Original GONG + GONG copy-paste & 0.43 \\
\bottomrule
\end{tabularx}
\end{table}

\subsection{Full pipeline with both synthetic sources}

The full YOLO + U-Net pipeline trained with original GONG observations, GONG copy-paste
composites, and CHASE-shape composites achieved PQ 0.4819 on the 91 development frames, compared
with 0.4719 for the original ensemble (Table~\ref{tab:6}). The gain is 0.0100 at the reported
four-decimal precision. It is a pipeline-level result, distinct from the standalone U-Net gain
in Table~\ref{tab:2}. TP, FP, and FN are not reported for the augmented pipeline.

The available augmented-pipeline score does not separate instance recovery, false-positive
suppression, and boundary overlap. Paired prediction sets and the exact checkpoint and fusion
configurations are required to analyze the gain and its uncertainty. Because training
composition and decoding policies differ, the change does not by itself isolate the contribution
of CHASE geometry.

\begin{table*}[t]
\centering
\caption{Full-pipeline development results on the same 91 GONG frames. The original ensemble uses multiple thresholds during fusion; the combined-source result is reported at threshold 0.50. An em dash denotes an unreported count. The original result is retained at the four-decimal precision used for this comparison.}
\label{tab:6}
\small
\renewcommand{\arraystretch}{1.12}
\setlength{\tabcolsep}{4pt}
\begin{tabularx}{\linewidth}{@{}Xrrrrr@{}}
\toprule
Training configuration & Threshold & PQ & TP & FP & FN \\
\midrule
Original GONG ensemble & Multiple & 0.4719 & 487 & 204 & 221 \\
Original GONG + GONG copy-paste + CHASE shapes & 0.50 & 0.4819 & --- & --- & --- \\
\bottomrule
\end{tabularx}
\end{table*}

\section{Discussion}

\subsection{Separating transferable geometry from intensity}

The CHASE procedure makes the separation between donor shape and target appearance explicit.
Binary support can be placed independently of the original CHASE intensity, and an empirical
GONG tuple supplies the decrement. The construction yields exact labels for the inserted
synthetic components and preserves background pixels outside them. Exact synthetic labels mean
exact correspondence with the chosen insertion support; they do not certify the physical
identity or completeness of the original automatic component.

The observed native operating point demonstrates that a model can be trained with this
construction without the catastrophic GONG deterioration seen in the separate continuation
study. It does not establish superiority over direct continuation under matched conditions: the
shape model starts from generic initialization, includes real GONG throughout the mixed
schedule, and receives different supervision and exposure. The continuation experiment starts
from solar-trained weights and removes GONG replay entirely. A comparison between their scores
cannot identify shape--appearance separation as the cause of the difference.

\subsection{What the current comparison does and does not test}

The available native baseline already contains GONG copy-paste. Consequently, Table~\ref{tab:2}
does not test augmentation against real-only training. The two synthetic arms also differ in
probability, object density, label rasterization, and runtime, as well as donor source. A
controlled study should compare real-only, GONG copy-paste, and CHASE shapes with the same
physical split, initialization policy, paired seeds, crop budget, and execution environment.
Both augmented arms should use a common mixture probability; crossing source with probabilities
0.20 and 0.50 would help separate source from dosage.

Matched training controls are required even when a common decoder is used. Repeated checkpoint
selection on the same development set can favor a transient point, particularly when only one
saved epoch exceeds the comparison score. A stronger assessment would freeze training and
selection rules before an untouched temporal cohort, report multiple seeds, and quantify the
uncertainty of paired per-frame differences. Until then, the results support feasibility and a
specific precision--recall trade-off rather than robust augmentation efficacy.

The full-pipeline development increase from 0.4719 to 0.4819 provides additional evidence at the
combined-system level. It does not show that the small standalone U-Net change explains the full
gain. The historical real-only YOLO result is also not a matched control for this configuration.
A complete comparison should preserve both pipeline prediction sets, apply a common decoding
policy or equal tuning budget, and quantify paired uncertainty before evaluating an untouched
temporal cohort.

\subsection{Boundary labels and repeated backgrounds}

The copy-paste route transfers genuine target texture but can carry local appearance context and
introduces a difference between the feathered intensity boundary and hard annotation support.
Permitted overlap also requires a consistent annotation convention. The CHASE route makes
support preservation explicit but inherits component errors from semantic pseudo-labels. Diercke
et al.~\cite{r1} likewise identify annotation and detection difficulties in solar archives;
their work motivates careful treatment of imperfect supervision, but does not validate these
generated instances or the present mask metric.

Using one quiet background is a further restriction. Its texture is repeated 462 or 616 times,
and any faint unlabeled feature remains part of the nominal background. Automated visibility
checks cannot prove that it is filament-free. Multiple provenance-controlled backgrounds would
test whether apparent benefits depend on foreground geometry or on a repeated texture.
Figures~\ref{fig:3} and~\ref{fig:4} provide actual observation/composite and reference-aligned
prediction examples, respectively. These selected panels do not establish how frequently faint
extensions, short filaments, and adjacent objects are represented faithfully; a systematic
paired analysis is still needed. The procedural Figure~\ref{fig:2} should be interpreted only as
a reproducible illustration.

\subsection{Generalization and reproducibility}

The fixed split groups identical observations but retains temporal relationships across frames.
It also serves repeatedly for method and checkpoint development. No independent instrument-level
generalization result or new challenge leaderboard gain is established here. The continuation
bootstrap conditions on fixed models and does not account for temporal correlation or seed
variation. For the small native difference, the absence of paired baseline masks precludes a new
paired uncertainty estimate or per-object correspondence analysis.

The code identifies the original native checkpoint and split files by checksum, but model
weights and complete run archives are external to the Git checkout. Reproducing the
Table~\ref{tab:2} operating points requires the selected checkpoints and their inference
configurations in addition to the code. The continuation comparison and its uncertainty summary
require the archived predictions and bootstrap configuration. The combined-source full-pipeline
result additionally requires its training mixture, checkpoints, prediction files, and exact
fusion settings. These records are necessary for independent reproduction of the reported
values.

\section{Conclusions}

We described two procedures for constructing labeled solar-filament training images:
target-instrument copy-paste and external mask geometry rendered with target-derived
attenuation. At a shared native decoder, the CHASE-shape configuration gives PQ 0.431117 versus
0.429419 for a GONG-copy-paste configuration, with fewer FP, more FN, and almost unchanged mean
matched IoU. Its selected decoding setting gives 0.431339. The small selected-point difference
is consistent with a precision--recall shift; it does not demonstrate better mask boundaries or
isolate the effect of the synthetic source.

A distinct CHASE-only continuation of both ensemble branches produces a large decline on GONG
under retained fusion rules. The combined-source full pipeline achieves development PQ 0.4819
versus 0.4719 for the original ensemble on the same 91-frame cohort. Early YOLO runs separately
recorded internal PQ 0.39 and 0.43 for real-only and GONG-copy-paste training. These historical
and full-pipeline results do not isolate the factors governing the standalone comparison.
Together, these observations motivate studying donor geometry, target appearance, replay, and
label granularity as separate factors. The next decisive evidence is a matched real-only and
two-source comparison with repeated seeds and an untouched temporal evaluation. The present
results establish an implemented synthesis route and its development behavior, while leaving
generalization, annotation efficiency, and computational benefit to be tested.

\section*{Data and code availability}

MAGFiLO and the challenge describe the target observations and evaluation setting~\cite{r2,r3}.
CHASE source masks are provided in the AHAS/MORDEN dataset release~\cite{r6,r7}. Implementation
and experiment documentation are available at
\url{https://github.com/ikjoo5/solar_filament}~\cite{r14}, reviewed commit \texttt{41ed6ea}. The
relevant entry points are \texttt{chase\_\allowbreak{}shape\_\allowbreak{}synthesis.py},
\texttt{chase\_\allowbreak{}shape\_\allowbreak{}training.py}, \texttt{solar\_\allowbreak{}a100\_\allowbreak{}train\_\allowbreak{}standalone.py}, and the frozen
submission fusion modules. Weights, experimental background pixels, complete predictions, and
run manifests are held outside the tracked repository; the code-only checkout is not a complete
numerical reproduction archive. The publicly reproducible renderer demonstration records its
donor, parameters, and source checksums separately from the training archive.

\section*{Acknowledgments}

We acknowledge the MAGFiLO annotators, the Solar Filament Segmentation Challenge organizers, and
the CHASE data providers. GONG observations are supplied by the NSO Integrated Synoptic Program.
The National Solar Observatory is operated by AURA under a cooperative agreement with the
National Science Foundation, with contributions from NOAA. Figure~\ref{fig:2} adapts CHASE mask
geometry from Hu et al.~\cite{r7} under CC BY 4.0. Figures~\ref{fig:3} and~\ref{fig:4} use
GONG/MAGFiLO observations and annotations; the synthesis background is identified in~\cite{r15},
and CHASE mask geometry is credited to~\cite{r7}. We acknowledge the developers of the
segmentation software used in this study.


\begin{thebibliography}{99}

\bibitem{r1}
Andrea Diercke, Robert Jarolim, Christoph Kuckein, Sergio J. Gonz\'alez Manrique,
Marco Ziener, Astrid M. Veronig, Carsten Denker, Werner P\"otzi,
Tatiana Podladchikova, and Alexei A. Pevtsov.
A universal method for solar filament detection from H$\alpha$ observations using
semi-supervised deep learning.
\emph{Astronomy \& Astrophysics}, 686:A213, 2024.
doi: \href{https://doi.org/10.1051/0004-6361/202348314}{10.1051/0004-6361/202348314}.

\bibitem{r2}
Azim Ahmadzadeh, Rohan Adhyapak, Kartik Chaurasiya, et al.
A dataset of manually annotated filaments from H-alpha observations.
\emph{Scientific Data}, 11:1031, 2024.
doi: \href{https://doi.org/10.1038/s41597-024-03876-y}{10.1038/\allowbreak s41597-024-03876-y}.

\bibitem{r3}
Azim Ahmadzadeh, Dustin J. Kempton, Qin Li, and Alexei A. Pevtsov.
Solar filament segmentation challenge 2026.
Kaggle, 2026.
URL \url{https://www.kaggle.com/competitions/filament-segmentation-2026}.

\bibitem{r4}
Golnaz Ghiasi, Yin Cui, Aravind Srinivas, Rui Qian, Tsung-Yi Lin,
Ekin D. Cubuk, Quoc V. Le, and Barret Zoph.
Simple copy-paste is a strong data augmentation method for instance segmentation.
In \emph{CVPR}, pages 2918--2928, 2021.

\bibitem{r5}
Chuan Li et al.
The chinese H$\alpha$ solar explorer (CHASE) mission: An overview.
\emph{Science China Physics, Mechanics \& Astronomy}, 65:289602, 2022.
doi: \href{https://doi.org/10.1007/s11433-022-1893-3}{10.1007/s11433-022-1893-3}.

\bibitem{r6}
J. R. Hu, Q. Hao, Z. Zheng, P. F. Chen, C. Li, and Y. Meng.
A modern ConvNet for solar filament detection, 2026.
URL \url{https://arxiv.org/abs/2607.24525}. arXiv:2607.24525v1.

\bibitem{r7}
Jiarui Hu, Qi Hao, Zheng Zhi, P. F. Chen, Chuan Li, and Yifan Meng.
Datasets for A Modern ConvNet for Solar Filament Detection.
Zenodo, 2026.
URL \url{https://doi.org/10.5281/zenodo.21582245}. Version 1.

\bibitem{r8}
Robert Jarolim, Astrid M. Veronig, Werner P\"otzi, and Tatiana Podladchikova.
A deep learning framework for instrument-to-instrument translation of solar observation data.
\emph{Nature Communications}, 16:3157, 2025.
doi: \href{https://doi.org/10.1038/s41467-025-58391-4}{10.1038/s41467-025-58391-4}.

\bibitem{r9}
Olaf Ronneberger, Philipp Fischer, and Thomas Brox.
U-Net: Convolutional networks for biomedical image segmentation.
In \emph{MICCAI}, pages 234--241, 2015.
doi: \href{https://doi.org/10.1007/978-3-319-24574-4_28}{10.1007/978-3-319-24574-4\_28}.

\bibitem{r10}
Zhuang Liu, Hanzi Mao, Chao-Yuan Wu, Christoph Feichtenhofer, Trevor Darrell,
and Saining Xie.
A ConvNet for the 2020s.
In \emph{CVPR}, pages 11976--11986, 2022.
URL \url{https://arxiv.org/abs/2201.03545}.

\bibitem{r11}
Alexander Kirillov, Kaiming He, Ross Girshick, Carsten Rother, and Piotr Doll\'ar.
Panoptic segmentation. In \emph{CVPR}, pages 9404--9413, 2019.
URL \url{https://arxiv.org/abs/1801.00868}.

\bibitem{r12}
Ultralytics. Ultralytics YOLO, 2026.
URL \url{https://github.com/ultralytics/ultralytics}.
Software dependency version 8.4.182.

\bibitem{r13}
Pavel Yakubovskiy. Segmentation models pytorch, 2020.
URL \url{https://github.com/qubvel-org/segmentation_models.pytorch}.
Software dependency version 0.5.0.

\bibitem{r14}
ikjoo5. Solar filament implementation and experiment records, 2026.
URL \url{https://github.com/ikjoo5/solar_filament}.
Reviewed commit 41ed6ea75f842a7a7f4b2d22d6dd8e109b70f2fc.

\bibitem{r15}
National Solar Observatory, Integrated Synoptic Program.
GONG H$\alpha$ full-disk observation, Learmonth, 2020 June 18 at 23:59:30 UT.
GONG image archive, 2020. File \href{https://gong2.nso.edu/HA/hag/202006/20200618/20200618235930Lh.jpg}{20200618235930Lh.jpg}.

\bibitem{r16}
Prairie View Solar Observatory, Prairie View A\&M University.
Solar observations 2020, june 18 entry.
\href{https://www.pvamu.edu/pvso/solar-observations-2020/?date=20200618}{Observing log}, 2020.

\bibitem{r17}
G. Castell\'o, M. Luna, and J. Terradas.
FILOS: a solar filament oscillation simulator, 2026.
\href{https://arxiv.org/abs/2611.22602}{arXiv:2611.22602}.

\end{thebibliography}
\end{document}